%% file: InvarianceOODG.tex
\documentclass{article}
\usepackage{spconf}
\usepackage{amsmath}
\usepackage{amsthm}
\usepackage{textcomp}
\usepackage{booktabs}
\usepackage{subfigure}

\usepackage{graphicx}
\usepackage{makecell}
\usepackage{float}
\usepackage{multirow}
\usepackage{amssymb}
\usepackage{stfloats}
\usepackage[normalem]{ulem}
\useunder{\uline}{\ul}{}

\title{InvariantOODG: Learning Invariant Features of Point Clouds for Out-of-Distribution Generalization}
\name{Zhimin Zhang, Xiang Gao, Wei Hu}
\address{Wangxuan Institute of Computer Technology, Peking University}

\begin{document}
%
\maketitle
\begin{abstract}
The convenience of 3D sensors has led to an increase in the use of 3D point clouds in various applications. However, the differences in acquisition devices or scenarios lead to divergence in the data distribution of point clouds, which requires good generalization of point cloud representation learning methods.
While most previous methods rely on domain adaptation, which involves fine-tuning pre-trained models on target domain data, this may not always be feasible in real-world scenarios where target domain data may be unavailable. To address this issue, we propose InvariantOODG, which learns invariability between point clouds with different distributions using a two-branch network to extract local-to-global features from original and augmented point clouds. 
Specifically, to enhance local feature learning of point clouds, we define a set of learnable anchor points that locate the most useful local regions and two types of transformations to augment the input point clouds.
The experimental results demonstrate the effectiveness of the proposed model on 3D domain generalization benchmarks.
\end{abstract}

\let\thefootnote\relax\footnotetext{Corresponding author: Wei Hu (forhuwei@pku.edu.cn). This work was supported by National Natural Science Foundation of China (61972009).}


\vspace{-0.15in} 
\section{Introduction}
\vspace{-0.05in}
\setlength {\parskip} {0cm}
\label{sec:intro}
\input{samples/01_introduction}

\vspace{-0.15in} 
\section{Method}
\vspace{-0.05in}
\label{sec:formulation}
\input{samples/03_method}

\vspace{-0.15in} 
\section{Experiments}
\vspace{-0.05in}
\label{sec:experiments}

\input{samples/05_experiments}

\vspace{-0.2in} 
\section{Conclusion}
\vspace{-0.05in}
\label{sec:conclusion}
\input{samples/06_conclusion}

\clearpage

\bibliographystyle{IEEEbib}
\bibliography{InvarianceOODG}

\end{document}

%% file: samples/01_introduction.tex
3D point clouds are discrete points sampled from 3D objects or scenes with various applications such as robotics, autonomous driving and telepresence. However, most existing point cloud learning models are trained on predictable synthetic datasets, while real-world point clouds vary significantly and often suffer from noise, missing regions, occlusions, \textit{etc}.
These gaps in data distribution require good generalization ability of point cloud learning models for practical applications.
Hence, it is essential to learn generalizable representations of point clouds that adapt to different data distributions and cope with the challenges of real-world data.

To address the generalization problem, many approaches have been proposed for domain adaptation \cite{long2015learning, ganin2016domain,  zhang2018aligning} and domain generalization \cite{yang2021causalvae, shen2022weakly} on images.
Domain adaptation involves adapting a pre-trained model from a source domain to a target domain, typically using labeled or unlabeled target data during training. By contrast, domain generalization aims to train a model on a source domain that can perform well on target domains where target data is not available during training. This is also known as out-of-distribution (OOD) generalization \cite{shen2021towards}.
While these methods have shown excellent performance on images \cite{yang2021causalvae,chang2020invariant}, few efforts have been made to learn generalizable representations for point clouds.
For point cloud domain adaptation, the most common approach is to learn cross-domain local features by aligning discriminant local regions \cite{qin2019pointdan}.
These methods usually sample some key points from the point cloud and then aggregate local regions. However, key points from different point clouds are often matched inaccurately.
In point cloud out-of-distribution generalization, the key idea is to design a range of data augmentation methods during the training to simulate geometric changes between the training and testing sets \cite{huang2021metasets, wei2022learning}.
Nevertheless, these methods heavily rely on the forms of data augmentation to simulate unseen data, which thus lacks flexibility.

To this end, we propose a novel approach to learn the invariability between point clouds with various distributions in both local key structures and global structure even in the case of incompleteness, which is referred to as InvariantOODG.
To achieve this goal, we design a two-branch network that can extract both local and global features from the original input point clouds and augmented ones, aiming to learn the invariance by minimizing the feature difference between the local and global features of the two branches.
Specifically, to achieve dynamic and accurate-matching learning of local features, we define a set of anchor points that can be used to identify specific local areas, which are learned from point cloud features, instead of being chosen from the original point cloud. 
Additionally, two transformation strategies are implemented to augment the point cloud and maintain the corresponding relationship of anchor points.
The learning of the network is constrained by several factors, including local and global feature loss, downstream task loss, and anchor point learning loss.

Our main contributions are summarized as follows.
\begin{itemize}
\setlength{\leftmargin}{0pt}
\setlength{\itemsep}{0pt}
\setlength{\topsep}{0pt}
\setlength {\parskip} {0cm}
\vspace{-0.1in}
  \item  We propose to learn invariant features of point clouds with various distributions in both local key structures and the global structure even in the case of incompleteness, aiming for out-of-distribution generalization of point clouds.
 \item We define anchor points and propose an anchor point learning module to facilitate dynamic and accurate matching of local regions for invariant feature learning. This module not only facilitates the learning of local features but also provides supplementary local information for incomplete inputs.
  \item Extensive experiments on 3D domain generalization benchmarks show the superiority of our method.
\end{itemize}

%% file: samples/03_method.tex
\vspace{-0.1in} 
\subsection{Problem Definition}
\vspace{-0.1in} 

The 3D point cloud OOD generalization task involves two datasets: the source domain dataset $\mathcal{D}_s=\left\{\mathcal{P}_s, \mathcal{Y}_s\right\}$ consisting of point clouds $\mathcal{P}_s=\{(\mathbf{X}_i^{(s)}\}_{i=1}^{P}$ and corresponding labels $\mathcal{Y}_s=\{\mathbf{y}_i^{(s)}\}_{i=1}^{P}$ with $P$ samples, and the target domain dataset $\mathcal{D}_t=\left\{\mathcal{P}_t,\mathcal{Y}_t\right\}$  with a different distribution\footnote{For better readability, we will drop the superscript $(s)$ and $(t)$ from now on.}.
The goal of point cloud OOD generalization is to train a highly generalized model $\mathcal{F}(\cdot)$ using the source domain dataset $\mathcal{D}_s$ that can achieve low classification errors $\varepsilon=\underset{(\mathbf{X},\mathbf{y})\sim\mathcal{D}_t}{\mathbb{E}}\left[\mathcal{F}(\mathbf{X})\ne\mathbf{y}\right]$ on the target domain dataset $\mathcal{D}_t$.
It is important to note that during the training phase, only the point cloud and the corresponding label $(\mathbf{X},\mathbf{y})$ of the source domain data $\mathcal{D}_s$ are available, while the samples of the target domain dataset are solely used for evaluating the generalization performance of the model $\mathcal{F}(\cdot)$.

\vspace{-0.15in} 
\subsection{Overview of the Model}
\vspace{-0.1in} 
\label{sec:model}
\input{samples/04_network_implementation}

\vspace{-0.15in} 
\subsection{Data Augmentation}
\vspace{-0.1in} 

We first define the parameterized and non-parameterized transformations to acquire the augmented point clouds for the subsequent invariant feature representation learning.
Given a point cloud $\mathbf{X}=\{\mathbf{x}_i\}_{i=1}^{N}\in\mathbb{R}^{N\times 3}$ with $N$ points, our goal is to obtain the transformed counterpart $\widetilde{\mathbf{X}}=\{\tilde{\mathbf{x}}_i\}_{i=1}^{N}\in\mathbb{R}^{N\times 3}$.

\textbf{Parameterized transformation.} Suppose we have a transformation distribution $\mathcal{T}_1=\left\{t_1\vert\theta\sim\mathbf{\Theta}\right\}$ with their parameters $\theta$ sampled from a distribution $\mathbf{\Theta}$, where $t_1(\cdot)$ denotes a transformation parameterized by $\theta$.
In this paper, we consider the parameterized transformation $t_1(\cdot)$ that can be represented by matrices, such as linear transformations, affine transformations (\textit{e.g.}, rotations, translations, \textit{etc.}).
The transformed point cloud can be written as
\begin{equation}
\small
\setlength\abovedisplayskip{3pt}
\setlength\belowdisplayskip{3pt}
  \widetilde{\mathbf{X}}=t_1(\mathbf{X})=\mathbf{X}\times T(\theta),
\end{equation}
where $T(\theta)\in\mathbb{R}^{3\times 3}$ denotes the transformation matrix.

\textbf{Non-parameterized transformation.} 
Other forms of transformations, such as randomly dropping points or resampling the original point cloud, can be used without explicit parameters like matrix transformations. 
We define this set of transformations are sampled from the distribution, \textit{i.e.}, $t_2\sim\mathcal{T}_2$, where $t_2(\cdot)$ is the non-parameterized transformation.
The transformed point cloud is defined as under the transformation $t_2(\cdot)$
\begin{equation}
\small
\setlength\abovedisplayskip{1pt}
\setlength\belowdisplayskip{1pt}
  \widetilde{\mathbf{X}}=t_2(\mathbf{X}).
\end{equation}

\vspace{-0.05in} 
In this paper, we consider the composition transformation of the two types of transformations $(t_2 \circ t_1)(\cdot)$ as the overall transformation to obtain the augmented point cloud, \textit{i.e.},
\begin{equation}
\small
\setlength\abovedisplayskip{1pt}
\setlength\belowdisplayskip{1pt}
\label{eq:transformation}
  \widetilde{\mathbf{X}}=(t_2 \circ t_1)(\mathbf{X})=t_2\left(t_1(\mathbf{X})\right).
\end{equation}
The parameterized transformation $t_1(\cdot)$ changes the position and pose of the point cloud $\mathbf{X}$, while non-parameterized transformation $t_2(\cdot)$ affects its density and distribution.

\vspace{-0.1in} 
\subsection{Local Invariance Learning}
\vspace{-0.1in}
\textbf{Dynamic Anchor Points Learning.}
We propose the dynamic anchor points learning module to learn a set of anchor points for local region detection.
We first extract point-level features $\widetilde{\mathbf{H}}\!=\!f(\widetilde{\mathbf{X}})$ from the augmented point cloud $\widetilde{\mathbf{X}}$,
where $\widetilde{\mathbf{H}}\!\in\!\mathbb{R}^{N \times D}$, $f: \mathbb{R}^{3}\!\mapsto\!\mathbb{R}^{D}$ denotes a point-level feature extractor, and $D$ is the hidden channels.
The feature representations $\widetilde{\mathbf{H}}$ will be fed into the proposed anchor point learning module to learn anchor points, \textit{i.e.},
\begin{equation}
\small
\setlength\abovedisplayskip{1pt}
\setlength\belowdisplayskip{1pt}
  \widetilde{\mathbf{A}}= \widetilde{\mathbf{A}}_0 + g_2(\mathbf{S}^{\top}\widetilde{\mathbf{H}}), \quad \text{with} \; \mathbf{S}=g_1(\widetilde{\mathbf{H}}),
  \label{eq:anchor_learning}
\end{equation}
where $\widetilde{\mathbf{A}}\!\in\!\mathbb{R}^{M \times 3}$ represents the learned $M$ anchor points from the augmented point cloud, $\widetilde{\mathbf{A}}_0$ is the initial point coordinates extracted by farthest point sampling (FPS) method, $\mathbf{S}\in\mathbb{R}^{N \times M}$ with $\sum_j s_{i,j}\!=\!1$ is a learned normalized selection matrix, $g_1:\mathbb{R}^{D}\!\mapsto\!\mathbb{R}^{M}$ and $g_2:\mathbb{R}^{D}\!\mapsto\!\mathbb{R}^{3}$ are two feature learning module.

Since the anchor point $\widetilde{\mathbf{A}}$ is extracted from the augmented point cloud, it has a similar pose to $\widetilde{\mathbf{X}}$.
Hence, we can directly obtain the anchor point with a similar pose to the original point cloud by applying the inverse transformation of parameterized transformation to $\widetilde{\mathbf{A}}$, \textit{i.e.},
\begin{equation}
\small
\setlength\abovedisplayskip{1pt}
\setlength\belowdisplayskip{1pt}
  \mathbf{A}=t_{1}^{-1}(\widetilde{\mathbf{A}}),
\end{equation}
where $t_{1}^{-1}(\cdot)$ is the inverse transformation of $t_1(\cdot)$.

We can easily prove that the proposed anchor point learning method is permutation invariant when the feature extractors $f(\cdot)$ and $g_1(\cdot)$ are permutation equivariant with Eq.~\ref{eq:invariance}. Thus, we can ensure the learning of anchor points is permutation invariant from the input point cloud, \textit{i.e.}, given two point clouds with the same point set but different point arrangements, they should extract the same anchor points.
\begin{equation}
\small
\setlength\abovedisplayskip{1pt}
\setlength\belowdisplayskip{1pt}
\label{eq:invariance}
\begin{split}
  & \widetilde{\mathbf{H}}^{\prime}=f(\mathbf{P}\mathbf{X})=\mathbf{P}\times f(\mathbf{X})=\mathbf{P}\mathbf{H}, \\
  & \mathbf{S}^{\prime}=g_1(\mathbf{P}\widetilde{\mathbf{H}})=\mathbf{P}\times g_1(\widetilde{\mathbf{H}})=\mathbf{P}\mathbf{S}.
\end{split}
\end{equation}

In practice, we choose the MLP-based architecture to implement $f(\cdot)$ and $g_1(\cdot)$ to ensure the permutation invariance.
In order to learn the anchor points, we choose the Chamfer distance between the extracted anchor points $\mathbf{A}$ and the original point cloud $\mathbf{X}$ as the loss function,
\begin{equation}
\setlength\abovedisplayskip{1pt}
\setlength\belowdisplayskip{1pt}
\small
  \mathcal{L}_{\text{CD}}=\frac{1}{M}\sum_{i=1}^{M}\min_{\mathbf{x}\in\mathbf{X}}\left\|\mathbf{a}_i-\mathbf{x}\right\|_2^2+\frac{1}{N}\sum_{i=1}^{N}\min_{\mathbf{a}\in\mathbf{A}}\left\|\mathbf{a}-\mathbf{x}_i\right\|_2^2.
  \label{eq:cd_loss}
\end{equation}

By minimizing the above loss function, the feature representations of point clouds and anchor points are jointly optimized, resulting in dynamically anchor point learning that evolves along with the features.

\textbf{Local Region Detection and Alignment.}
Having obtained the anchor points $\mathbf{A}$ and $\widetilde{\mathbf{A}}$, we next use them to detect and align the local regions for both the original and augmented point clouds for local feature extraction.
Using anchor points $\mathbf{A}$, we detect local regions in the original point cloud. 
For each anchor point, we use the following approach to obtain several point sets that serve as local regions:
\begin{equation}\small
\setlength\abovedisplayskip{1pt}
\setlength\belowdisplayskip{1pt}
  \mathcal{N}_i=\left\{j\;\big\vert\;\|\mathbf{a}_i-\mathbf{x}_j\|_2^2<r, \mathbf{x}_j\in\mathbf{X}\right\}, \forall \mathbf{a}_i\in\mathbf{A},
\end{equation}
where $\mathcal{N}_i$ denotes the local point set by the index that is located by anchor point $\mathbf{a}_i$, and $r$ is a parameter.
Next, we use the same method to obtain the local regions of the augmented point cloud.
We detect the local regions in $\widetilde{\mathbf{X}}$ by $\widetilde{\mathbf{A}}$ as
\begin{equation}\small
\setlength\abovedisplayskip{1pt}
\setlength\belowdisplayskip{1pt}
  \widetilde{\mathcal{N}}_i=\left\{j\;\big\vert\;\|\tilde{\mathbf{a}}_i-\tilde{\mathbf{x}}_j\|_2^2<r, \tilde{\mathbf{x}}_j\in\widetilde{\mathbf{X}}\right\}, \forall \tilde{\mathbf{a}}_i\in\widetilde{\mathbf{A}}.
\end{equation}
Since the transformation relationship between $A$ and $\widetilde{\mathbf{A}}$ is consistent with that between $X$ and $\widetilde{\mathbf{X}}$, we can obtain accurate local region matching between the two point clouds.
Hence, we can detect and align local regions in both the original and augmented point clouds, denoted as $\left\{\mathcal{N}_i,\widetilde{\mathcal{N}}_i\right\}$, $i=1,...,M$.

\textbf{Local Feature Aggregation.}
After obtaining these local regions $\left\{\mathcal{N}_i,\widetilde{\mathcal{N}}_i\right\}$, we next aggregate their feature representations for local invariant learning. 
Given the anchor points $\{\mathbf{A},\widetilde{\mathbf{A}}\}$ and the corresponding point clouds $\{\mathbf{X},\widetilde{\mathbf{X}}\}$, we first extract the feature representations of these anchor points, denoted as $\mathbf{H}^{(\text{A})}\in\mathbb{R}^{M \times D}$.
We then leverage a message passing layer to aggregate the local features located by the anchor points, \textit{i.e.},
\begin{equation}\small
\setlength\abovedisplayskip{1pt}
\setlength\belowdisplayskip{1pt} \mathbf{f}_{i}=w_{i,i}\mathbf{h}^{(\text{A})}_{i}+\sum_{j\in\mathcal{N}_i\setminus i}w_{i,j}\mathbf{h}_j, \quad i=1,...,M,
\end{equation}
where $\mathbf{h}_j\!\in\!\mathbf{H}$ and $\mathbf{h}^{(\text{A})}_{i}\!\in\!\mathbf{H}^{(\text{A})}$, $\mathbf{f}_i\!\in\!\mathbf{F}$ with $\mathbf{f}_i\!\in\!\mathbb{R}^{D}$ denotes the aggregated local features, and $w_{i,j}$ is the edge weight between node $i$ and $j$ where we use the graph attention network \cite{velickovic2017gat} to dynamically learn these edge weights.
Similarly, we can obtain the local features $\widetilde{\mathbf{F}}$ of the augmented point cloud.

Further, we extract more layers of anchor points to capture local features at different scales.
We first denote the point-level features as $\{\mathbf{H}^{(l)},\widetilde{\mathbf{H}}^{(l)}\}$ with the feature extractor $f^{(l)}(\cdot)$, the anchor points as $\{\mathbf{A}^{(l)},\widetilde{\mathbf{A}}^{(l)}\}$, and the aligned local features as $\{\mathbf{F}^{(l)},\widetilde{\mathbf{F}}^{(l)}\}$ at the $l$th layer, respectively.
Thus, we can extract the point-level feature representations at the $(l+1)$th layer by
\begin{equation}\small
\setlength\abovedisplayskip{1pt}
\setlength\belowdisplayskip{1pt}
  \mathbf{H}^{(l+1)}=f^{(l+1)}\left([\mathbf{H}^{(l)}\| \mathbf{F}^{(l)}]\right),
\end{equation}
where $[\cdot\|\cdot]$ is a concatenation operator, $f^{(l+1)}(\cdot)$ denotes the feature extractor.
Hence, by applying the operation in Eq.~(\ref{eq:anchor_learning}) to the point-level features $\{\mathbf{H}^{(l+1)},\widetilde{\mathbf{H}}^{(l+1)}\}$, we can obtain the anchor points $\{\mathbf{A}^{(l+1)},\widetilde{\mathbf{A}}^{(l+1)}\}$ at the $(l+1)$th layer for further local feature learning at a different scale.

Thus, by minimizing the following objective, we achieve the local invariant feature learning,
\begin{equation}\small
\setlength\abovedisplayskip{1pt}
\setlength\belowdisplayskip{1pt}
  \mathcal{L}_{\text{local}}=\sum_{l=1}^{L}\|\mathbf{F}^{(l)}-\widetilde{\mathbf{F}}^{(l)}\|_F^2,
  \label{eq:local_inv}
\end{equation}
where $\mathbf{F}^{(l)}$ and $\widetilde{\mathbf{F}}^{(l)}$ are aligned local features, and $L$ is the number of layers we deployed for local invariance learning.
\vspace{-0.15in} 
\subsection{Global Invariance Learning}
\vspace{-0.1in}
We further learn global invariant feature representations.
After obtaining the point-level feature representations $\textstyle{\!\{\mathbf{H}^{(L)}\!,\widetilde{\mathbf{H}}^{(L)\!}\}}$ of the last layer, we first use a pooling operator to obtain the global descriptor of the point cloud, \textit{i.e.},
\begin{equation}\small
\setlength\abovedisplayskip{1pt}
\setlength\belowdisplayskip{1pt}
  \mathbf{g}=\mathrm{pooling}(\mathbf{H}^{(L)}), \; \tilde{\mathbf{g}}=\mathrm{pooling}(\widetilde{\mathbf{H}}^{(L)}),
\end{equation}
where $\{\mathbf{g},\tilde{\mathbf{g}}\}$ are the global feature vectors of the original and augmented point clouds.

Then, we minimize the following objective for global invariance learning,
\begin{equation}\small
  \mathcal{L}_{\text{global}}=\|\mathbf{g}-\tilde{\mathbf{g}}\|^2_2.
  \label{eq:global_inv}
\end{equation}

\vspace{-0.2in} 
\subsection{The Overall Loss Function}
\vspace{-0.1in}
In summary, the entire network is trained end-to-end by minimizing the loss:
\begin{equation}\small
\setlength\abovedisplayskip{1pt}
\setlength\belowdisplayskip{1pt}
\mathcal{L}=\mathcal{L}_{\text{task}}+\alpha\mathcal{L}_{\text{CD}}+\beta\mathcal{L}_{\text{local}}+\gamma\mathcal{L}_{\text{global}},
  \label{eq:overall_loss}
\end{equation}
where $\alpha$, $\beta$ and $\gamma$ are parameters, $\mathcal{L}_{\text{task}}$ denotes the loss function for the downstream tasks, \textit{e.g.}, the cross-entropy loss
\begin{equation}\small
\setlength\abovedisplayskip{1pt}
\setlength\belowdisplayskip{1pt}
  \mathcal{L}_{\text{task}}=-\frac{1}{P}\sum_{i=1}^{P}\sum_{j=1}^{C}y_{i,j}\log\hat{y}_{i,j},
\end{equation}
where $P$ is the number of training samples in the source domain, and $C$ is the number of classes, $\mathbf{y}_i\in\mathbb{R}^{C}$ and $\hat{\mathbf{y}}_i\in\mathbb{R}^{C}$ are the ground-truth and predicted labels, respectively.

%% file: samples/04_network_implementation.tex
\begin{figure*}[t]
  \centering
  \includegraphics[width=\textwidth]{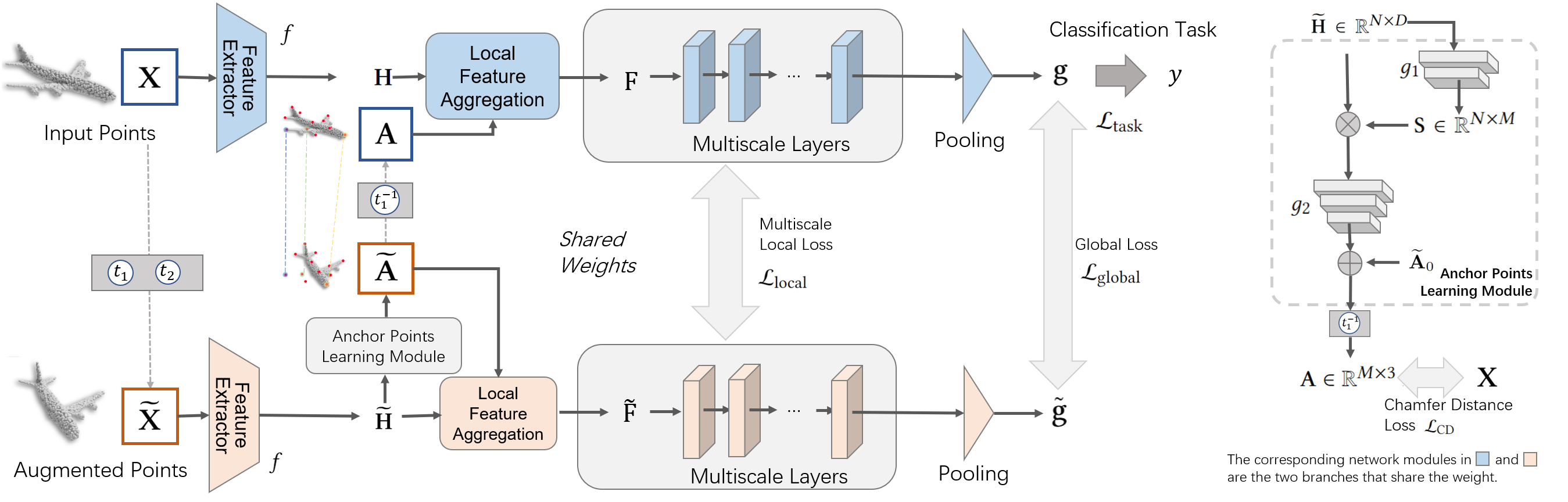}
  \caption{An illustration of the overall framework.}
  \label{fig:network}
  \vspace{-0.15in}
\end{figure*}
The proposed InvariantOODG framework is shown in Fig.~\ref{fig:network}. 
The original point cloud $\mathbf{X}$ is first transformed into the augmented point cloud $\widetilde{\mathbf{X}}$ using parameterized and non-parameterized transformations $t_1(\cdot)$ and $t_2(\cdot)$.
Both point clouds are fed into a two-branch feature extractor $f(\cdot)$ with shared weights.  The learned features $\{\mathbf{H},\widetilde{\mathbf{H}}\}$ then undergo anchor point learning and feature aggregation.
The anchor points learning module consists of two feature learning modules $g_1(\cdot)$ and $g_2(\cdot)$, and the local features are aggregated using a graph attention layer.
Finally, we employ a max-pooling layer to obtain the global features for further global invariance learning.
Next, we will discuss the details.

%% file: samples/05_experiments.tex
\begin{table*}[!ht]
\small
\centering
\caption{Classification accuracy (\%) on the PointDA datasets. The underlined text indicates that that there is a domain gap of simulation and reality between these two datasets.}
\begin{tabular}{ccccccccccc}
\hline
\multirow{2}{*}{\textbf{Methods}} & \multicolumn{2}{c}{\textbf{Sim-to-Real}} & \multicolumn{6}{c}{\textbf{PointDA}}                                    & \multirow{2}{*}{\textbf{Average}} & \multirow{2}{*}{\textbf{\makecell{Average \\ on Underlined}}} \\
                                  & {\ul M11\textrightarrow SO11}      & {\ul S9\textrightarrow SO9}     & {\ul M\textrightarrow S*} & {\ul S\textrightarrow S*}   & {\ul S*\textrightarrow M} & {\ul S*\textrightarrow S}   & M\textrightarrow S & S\textrightarrow M &                                   &                                          \\ \hline
MetaSets                          & 60.3                 & 51.8              & 52.3        & 42.1          & 69.8        & 69.5          & \textbf{86.0} & 67.3 & 62.4                              & 57.6                                     \\
PDG                               & 67.6                 & 57.3              & \textbf{57.9}        & 50.0          & \textbf{70.3}        & 66.3          & 85.6 &\textbf{ 73.1} & 66.0                              & 61.6                                     \\
\textbf{InvarianceOODG}           & \textbf{69.8}        & \textbf{59.8}     & 56.4        & \textbf{57.6} & 69.5        & \textbf{73.5} & 83.7 & 71.7 & \textbf{67.8}                     & \textbf{64.4}                            \\ \hline
\end{tabular}
\label{tab:PointDA}
\end{table*}

\subsection{Experiments Settings}
\vspace{-0.1in}
\textbf{Baseline}
We compare our InvarianceOODG with the baseline of point cloud OOD generalization methods such as MetaSets \cite{huang2021metasets} and PDG \cite{wei2022learning}.

\textbf{DataSets} 
We adopt two datasets for classification experiments following PDG\cite{wei2022learning}. 
Sim-to-Real dataset contains 11 common categories between ModelNet\cite{modelnet} and ScanObjectNN\cite{scanobjectnn}, denoted as $M11$ and $SO11$, as well as 9 common categories between ShapeNet and ScanObjectNN, denoted as $S9$ and $SO9$.
PointDA dataset is a commonly utilized point cloud domain adaptation benchmark that includes shapes from ten shared classes taken from ModelNet \cite{modelnet} (denoted as $M$), ShapeNet \cite{chang2015shapenet} (denoted as $S$), and ScanNet \cite{dai2017scannet} (denoted as $S^*$).
Among them, ModelNet($M9$,$M11$ and $M$) and ShapeNet($S9$ and $S$) are simulation datasets, while ScanObjectNN($SO9$ and $SO11$) and ScanNet($S^*$) are real datasets. $A$\textrightarrow$B$ indicate a generalization task from source dataset $A$ to unseen target dataset $B$.

\textbf{Implementation Details}
In our model, we adapt PointNet \cite{qi2017pointnet} as our backbone network of feature extractors $f^{(l)}(\cdot)$, $l=1,...,L$, \textit{i.e.}, we decompose PointNet into several parts and take each part of the network as $f^{(l)}(\cdot)$.
During training, the Adam \cite{kingma2014adam} optimizer is deployed with a batch size of 16.
The initial learning rate is 0.001 and weight decay is 0.0001, and the learning rate is decayed every 20 steps for 200 training epochs.
The three parameters $\alpha$, $\beta$ and $\gamma$ in Eq.~(\ref{eq:overall_loss}) are set to 1.0, respectively. We set the number of anchor points to 256 and the number of multiple layers to 2.

\vspace{-0.1in}
\subsection{Classification Results}
\vspace{-0.1in}
Table~\ref{tab:PointDA} presents the performance of our method on the Sim-to-real dataset and PointDA dataset. The underlined text indicates that that there is a domain gap of simulation and reality between these two datasets.
As we can see, the proposed method achieves the best OOD classification accuracies of $69.8\%$ and $59.8\%$ under the M11 \textrightarrow SO11 and S9 \textrightarrow SO9 tasks, respectively.
On the PointDA dataset, our method achieves state-of-the-art performance for most of the tasks and comparable performance to state-of-the-art models for other tasks.
Our method achieves the best result on average accuracy, which is more obvious on datasets with simulation and reality domain gaps, demonstrating the generalization effect of our method.

\vspace{-0.1in}
\subsection{Ablation Study}
\vspace{-0.1in}
We further performed ablation experiments on the main components of InvarianceOODG for the $M11$\textrightarrow$SO11$ task, as shown in Table \ref{tab:ablation}. For InvarianceOODG w/o anchor points, we use the farthest point sampling (FPS) method to extract the same number of points as anchor points for further representation learning.

\vspace{-0.2in}
\begin{table}[htbp]
\centering
\footnotesize
\centering
\caption{Ablation studies on the main components.}
\begin{tabular}{c p{1.5cm} p{1.5cm} p{1.5cm} c}
\hline
\multirow{2}{*}{\textbf{PointNet}} & \multicolumn{4}{c}{\textbf{InvarianceOODG}}                               \\ \cline{2-5} 
                                   & w/o Anchor Points & w/o Local Invariance & w/o Global Invariance & w/ all \\ \hline
59.8                               & 67.5              & 65.9                 & 67.5                  & 69.8   \\ \hline
\end{tabular}
\label{tab:ablation}
\end{table}

%% file: samples/06_conclusion.tex
In this paper, we present a novel method named InvariantOODG to learn the invariability between point clouds with various distributions, in terms of both local key structures and global structure even in the case of incompleteness.
In particular, we propose a two-branch framework to learn invariant features of point clouds before and after augmentation. 
To match local regions more accurately and more emphatically, we further propose a dynamic anchor point learning module to optimize the learning of local features, which provides local representation for even incomplete point clouds.
In the experiments on sim-to-real and PointDA datasets, we have demonstrated the superiority of our method over the state-of-the-art methods for out-of-distribution generalization.